\documentclass[runningheads]{llncs}
\usepackage[T1]{fontenc}
\usepackage{wrapfig} % added by -huub
\usepackage{graphicx}
\usepackage[utf8]{inputenc}
\usepackage{url}
\usepackage{enumitem}
\usepackage{algorithm}
\usepackage[noend]{algpseudocode}
\usepackage{listings}
\usepackage{xcolor}
\usepackage{mdframed}

\usepackage{mathtools} % cases
\usepackage{booktabs} % addlinespace
\usepackage{graphicx} % Required for inserting images
\usepackage{subfigure}
\usepackage{caption}
\usepackage{inconsolata}

\usepackage{tikz}
\usetikzlibrary{calc, shapes} % for coordinate calculations
\usetikzlibrary{positioning}

\tikzset{
  gamenode/.style={
    shape=ellipse,
    draw,
    thin,        % 'semithick', 'thick'
    inner sep=1pt,
    font=\tiny,
    minimum height=4mm,
    minimum width=6mm,
  },
  circlenode/.style={
    shape=ellipse,
    draw,
    thin,        % 'semithick', 'thick'
    inner sep=1pt,
    font=\tiny,
    minimum height=4mm,
    minimum width=4mm,
  }, 
  whiteplayer/.style={fill=white},
  blackplayer/.style={fill=gray!50},
  doubleborder/.style={double},
  dashednode/.style={
    dash pattern=on 1pt off 1pt   % finer dashes
  },
  gamelabel/.style={
    draw,
    rectangle,
    fill=white,
    inner sep=1pt,
    font=\tiny,
    anchor=north east,
    shift={(-0.34,0.1)}
  },
  gameedge/.style={->, thin, >=stealth},
  gamedashedge/.style={
    ->,
    thin,
    >=stealth,
    dash pattern=on 1pt off 1pt  % finer dashed line
  }
}

\newenvironment{algphase}[2]
{
\begin{mdframed}[
  backgroundcolor=#1,
  roundcorner=4pt,
  innertopmargin=2pt,
  innerbottommargin=2pt
]
\begin{algorithmic}
}
{
\end{algorithmic}
\end{mdframed}
}

\newfloat{lstfloat}{htbp}{lop}
\floatname{lstfloat}{Listing}

\lstdefinelanguage{Dafny}{
  keywords={method, returns, ensures, if, else, int, type, datatype, const, var, invariant, assert, return, ghost, reveal, decreases, class, requires, break, modifies, then, predicate, forall, exists, lemma, for, function, reads},
  sensitive=true,
  comment=[l]{//},
  morecomment=[s]{/*}{*/},
  morestring=[b]",
}

\definecolor{codegray}{rgb}{0.5,0.5,0.5}
\definecolor{codered}{rgb}{0.8,0,0}
\definecolor{keywordblue}{rgb}{0.0, 0.2, 0.6} % A professional Navy Blue
\definecolor{lemmateal}{rgb}{0.0, 0.4, 0.4}   % A distinct but muted Teal
\definecolor{vividOrange}{RGB}{200, 80, 0}

\newcommand{\attr}[1]{\ensuremath{\textit{#1}}} % attributes
\newcommand{\func}[1]{\ensuremath{\textsf{#1}}} % functions
\newcommand{\var}[1]{\ensuremath{\textit{#1}}} % variables
\newcommand{\set}[1]{\ensuremath{\{ #1 \} }} % sets
\newcommand{\truncate}[2]{\ensuremath{\left \lfloor{#1}\right \rfloor_{#2}}} % truncate game trees

\begin{document}
\title{Formal Verification of Minimax Algorithms}

\author{
  Wieger Wesselink \inst{1}\orcidID{0009-0001-6746-5115} \and \\
  Kees Huizing \inst{1}\orcidID{0000-0002-4713-7253} \and \\
  Huub van de Wetering \inst{1}\orcidID{0000-0002-0517-1322}
}
\authorrunning{W. Wesselink et al.}

\institute{Eindhoven University of Technology, The Netherlands \\
  \email{\{j.w.wesselink,h.v.d.wetering,c.huizing\}@tue.nl}
}

\maketitle

\begin{abstract}
Minimax-based search algorithms with alpha-beta pruning and transposition tables are a central component of classical game-playing engines and remain widely used in practice. Despite their widespread use, these algorithms are subtle, highly optimized, and notoriously difficult to reason about, making non-obvious errors hard to detect by testing alone.
Using the Dafny verification system, we formally verify a range of minimax search algorithms, including variants with alpha-beta pruning and transposition tables. For depth-limited search with transposition tables, we introduce a witness-based correctness criterion that captures when returned values can be justified by an explicit game-tree expansion. We apply this criterion to two practical variants of depth-limited negamax with alpha-beta pruning and transposition tables: for one variant, we obtain a fully mechanized correctness proof, while for the other we construct a concrete counterexample demonstrating a violation of the proposed correctness notion. All verification artifacts, including Dafny proofs and executable Python implementations, are publicly available.
\end{abstract}

\section{Introduction}
Game tree search algorithms, such as minimax and its optimization via alpha-beta pruning, compute optimal moves by recursively evaluating all possible outcomes under the assumption of perfect play. These algorithms assign values to game states based on the principle that one player seeks to maximize the evaluation, while the opponent seeks to minimize it.
The alpha-beta pruning technique, which may significantly improve the efficiency of minimax by pruning branches that cannot affect the final decision, was analyzed in detail by Knuth and Moore in their seminal work~\cite{KnuthM75}. Various theoretical frameworks have been applied to minimax and related algorithms, including lattice theory~\cite{BirdH87} and AND/OR tree representations~\cite{Stockman79}.
Several variations of alpha-beta pruning exist. In~\cite{KnuthM75}, the \emph{fail-hard} variant of alpha-beta pruning is discussed, while Fishburn introduced \emph{fail-soft} variants in~\cite{Fishburn80,Fishburn83}.
More recently, Nipkow et al.\ have verified implementations of several minimax variants in a functional programming setting using the Isabelle proof assistant~\cite{nipkow2024functional}. Their work includes a verified transposition-table-based algorithm, but only in the setting of depth-unlimited search.
This setting avoids many of the subtleties that arise in depth-limited search, where partially explored subtrees and varying depths complicate the formulation and verification of correctness criteria for transposition table usage.
In this paper, we focus on the formal verification of minimax and negamax~\cite{Stockman79} algorithms in an imperative setting using Dafny, a programming language and verifier designed for writing provably correct programs~\cite{Leino10}.
Dafny is grounded in Floyd-Hoare logic and supports automated reasoning through constructs such as preconditions, postconditions, assertions, loop invariants, and lemmas.

The central challenge we address is the formalization of correctness for depth-limited search with transposition tables, where cached results obtained at different depths and under different search windows break the simple tree-traversal semantics typically used to justify minimax and alpha-beta pruning. In this setting, correctness can no longer be explained solely in terms of a single execution tree. To address this, we introduce a flexible, witness-based correctness criterion that can account for partially explored subtrees and reused results from transposition tables. Using this framework, we verify both standard depth-limited minimax/negamax and practical extensions with transposition tables. These algorithms are widely used in game-playing engines and are known to be subtle and error-prone; even small implementation differences can lead to unexpected behavior. Our work demonstrates how formal verification can provide a robust foundation for reasoning about their correctness in practice.

\begin{wrapfigure}[12]{R}{0.35\textwidth}
    \vspace{-75pt}
    \centering  % [trim={left bottom right top},clip]
    \includegraphics[trim={1.4cm 1.4cm 0.5cm 0.5cm},clip, width=0.38\textwidth]{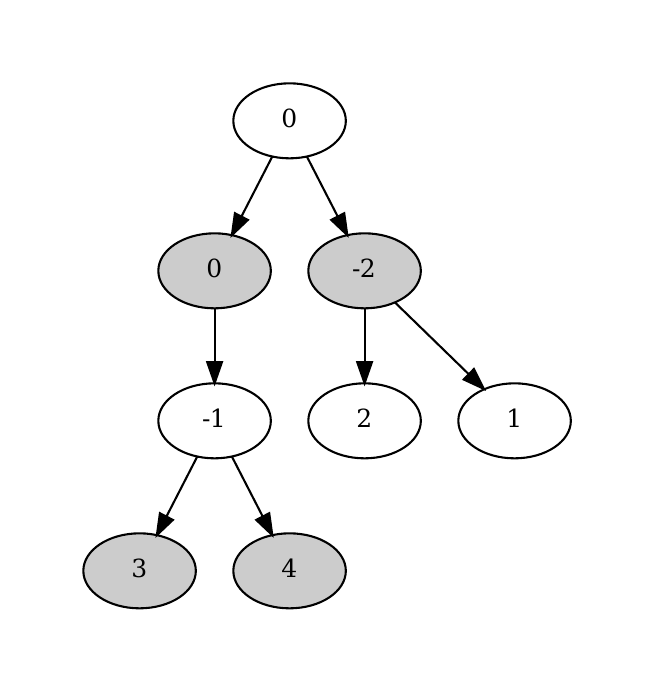}
    \caption{Game tree with white maximizer nodes and gray minimizer nodes, labeled with evaluation values.}
    \label{fig:gametree}
\end{wrapfigure}
\section{Definitions} \label{sec:definitions}
A game tree is a rooted directed acyclic graph representing the possible evolutions of a two-player game. Each node corresponds to a game position, and each edge to a legal move. Nodes are colored by the player to move: white for the maximizer and gray for the minimizer, as illustrated in Figure~\ref{fig:gametree}.
Formally, a node is defined as a record with three attributes:
\[
\text{Node} = \textbf{record} \left\{
\begin{array}{ll}
    \attr{eval} & \text{: evaluation value} \\
    \attr{color} & \text{: player to move ($\pm 1$)} \\
    \attr{children} & \text{: ordered child list}
\end{array}
\right.
\]
The \attr{eval} field encodes an integer value of a game state: the maximizer (\emph{color}=1) favors higher, while the minimizer (\emph{color}=-1) favors lower values. Internal nodes use heuristic evaluations, while leaves reflect game-theoretic values.
The $\attr{children}$ list defines the successors of a node $u$ in a fixed order, inducing edges $(u,v)$ iff $v \in u.\attr{children}$. A fixed order is not strictly necessary, but it is commonly used and enables deterministic execution.
We assume acyclicity, as repetition rules in games like chess and draughts preclude infinite loops in practice. Nodes abstract from concrete game states: equality is structural, requiring identical \attr{eval}, \attr{color}, and isomorphic subtrees. This suffices for search correctness while avoiding full state representation.
Since a node $u$ implicitly defines the subtree rooted at $u$, we use the terms node and game tree interchangeably. For a game tree $u$ we recursively define $\func{minimax}(u)$ as
\begin{align}
    \begin{cases*}
      u.\attr{eval} & \text{if $|u.\attr{children}| = 0$} \\
      \min \set{ \func{minimax}(v) \mid v \in u.\attr{children}} & \text{if $|u.\attr{children}| > 0$ and $u.\attr{color} = -1$} \\
      \max \set{ \func{minimax}(v) \mid v \in u.\attr{children}} & \text{if $|u.\attr{children}| > 0$ and $u.\attr{color} = 1$} \\
    \end{cases*}
\end{align}
and $\func{negamax}(u)$ as
\begin{align}
    \begin{cases*}
      u.\attr{color} \cdot u.\attr{eval} & \text{if $|u.\attr{children}| = 0$} \\
      \max \set{ - \func{negamax}(v) \mid v \in u.\attr{children}} & \text{if  $|u.\attr{children}| > 0$}. \\
    \end{cases*}
\end{align}
For minimax, we allow the same player to move consecutively. For negamax, the game must be \emph{turn-based}, i.e., for each node $u$ the following holds:
\begin{align}
  \forall v \in u.\attr{children} : v.\attr{color} = - u.\attr{color}.
\end{align}
Note that the two functions are the same up to a sign inversion.
\begin{equation}
  \func{negamax}(u) = u.\attr{color} \cdot \func{minimax}(u).
\end{equation}
In depth-limited search, we work with truncated game trees.
For a game tree $u$ with root node $r$, the \emph{depth-$d$ truncation} $\truncate{u}{d}$ is the subtree consisting of all nodes whose path from $r$ has length at most $d$, along with all edges connecting them. The root $r$ is at distance 0, and nodes at distance greater than $d$ are excluded.

\subsection{Alpha-beta Pruning}
Alpha-beta pruning is an optimization of minimax search that avoids exploring parts of the game tree without affecting the final decision. The algorithm maintains a window $(\alpha, \beta)$ of values that are relevant to the outcome. If the exact minimax (or negamax) value lies
strictly within this window, it must be computed precisely; if it lies outside the window, any value that preserves the appropriate bound is sufficient. This allows the search to terminate early in branches that are irrelevant to the root decision.
To formalize this behavior, we introduce several predicates used in the verification of alpha-beta pruning:
\begin{align}
  &\func{is-ab-result}(x, e, \alpha, \beta) =
       (e \leq x \leq \alpha)
       \;\lor\;
       (\alpha < e = x < \beta)
       \;\lor\;
       (\beta \leq x \leq e), \\[1ex]
  & \func{is-negamax-ab-result}(x, u, \alpha, \beta) =
     \func{is-ab-result}(x, \func{negamax}(u), \alpha, \beta)
     \label{eq:is-negamax-ab-result}
\end{align}
where the predicate \func{is-ab-result}, adapted from Fishburn~\cite{Fishburn83}, specifies when
a value $x$ is an acceptable result for a true minimax or negamax value $e$ under the window
$(\alpha,\beta)$. The three cases correspond to returning an upper bound ($e \leq x$), an exact value ($e = x$), or a lower bound ($x \leq e$), respectively.
To support reasoning about partial computations, we also define
\begin{equation}
  \func{is-partial-negamax-ab-result}(x, u, i, \alpha, \beta) =
     \func{is-ab-result}(x, \func{pnm}(u, i), \alpha, \beta),
  \label{eq:is-partial-negamax-ab-result}
\end{equation}
where $\func{pnm}(u, i) = \max \set{ -\func{negamax}(u.\attr{children}[j]) \mid 0 \leq j < i}$
denotes the partial negamax value obtained after evaluating the first $i$ children of $u$. This notion is used to express loop invariants and intermediate correctness conditions in the verification.

\subsection{Transposition Tables}
A transposition table (TT) is a cache used to avoid repeated computations in game-tree search by storing results of previously visited states. Formally, it represents a partial mapping from states (nodes) to search results. We define a table entry as follows:
\[
\begin{array}{@{}l}
\text{TableEntry} = \textbf{record} \\[0.3em]
~~~~~~ \left\{
  \begin{array}{@{}ll}
    \attr{value} & \text{ estimated minimax value,} \\ 
    \attr{depth} & \text{ search depth at evaluation time,} \\
    \attr{flag}  & \text{ value classification (see below).}
  \end{array}
\right.
\end{array}
\]
The $\attr{flag}$ attribute identifies the relationship between the stored $\attr{value}$ and an exact minimax or negamax value of the state:
\[
\begin{cases}
    \func{exact} ~~(\func{EX}) & \text{$value$ is an exact value,} \\
    \func{lowerbound} ~~(\func{LB}) & \text{$value$ is a lower bound,} \\
    \func{upperbound} ~~(\func{UB}) & \text{$value$ is an upper bound.}
\end{cases}
\]
Despite their widespread use, the formal semantics of transposition tables is typically underspecified. While the flag indicates how the stored value relates to an exact minimax value, it does not specify \emph{with respect to which game tree} this exact value is defined. In particular, table entries may be produced by searches performed at different depths, under different alpha-beta windows, and in different parts of the game tree. As a result, the meaning of a table entry cannot be understood in isolation: its validity depends implicitly on a search context that is not recorded in the table itself. This lack of a precise semantic foundation complicates formal reasoning about the correctness of transposition-table-based search.

In this work, we intentionally restrict the contents of the transposition table to values, depths, and flags, and do not store best moves. Best moves are commonly used to reorder successor nodes and thereby improve search efficiency, but they do not affect the correctness of returned values. Since our focus is on establishing value correctness rather than on performance characteristics, we exclude best-move information from the formal model.

%%%%%%%%%%%%%%%%%%%%%%%%%%%%%%%%%%%%%%%%%%%%%%%%%%%%
\algrenewcommand\algorithmicindent{0.7em}
\begin{figure}[tbh]
%\caption{Minimax and Negamax}
\centering
\begin{minipage}[t]{0.48\textwidth}
    \begin{algorithm}[H]
    \small
    \caption{Minimax}
    \label{alg:minimax}
    \vspace*{1ex}
    {\textbf {Input}:} A game tree $u$.\\
    {\textbf {Output}:} $\func{minimax}(u)$.\\  [1ex]
    {\textsc{minimax}($u$):}
    \begin{algorithmic}[1]
    \If {$|u.\attr{children}| = 0$}
      \State \Return $u.\attr{eval}$
    \ElsIf{ $u.\attr{color} = -1$ }
      \State $\var{value} := \infty$
      \For{$v \in u.\attr{children}$}
        \State $\var{value} := \min(\var{value}, \textsc{minimax}(v))$
      \EndFor
      \State \Return $\var{value}$
    \Else
      \State $\var{value} := -\infty$
      \For{$v \in u.\attr{children}$}
        \State $\var{value} := \max(\var{value}, \textsc{minimax}(v))$
      \EndFor
      \State \Return $\var{value}$
    \EndIf
    \end{algorithmic}
    \end{algorithm}
\end{minipage}
\hfill
\begin{minipage}[t]{0.5\textwidth}
    \begin{algorithm}[H]
    \small
    \caption{Negamax}
    \label{alg:negamax}
    \vspace*{1ex}
    {\textbf {Input}:} A game tree $u$.\\
    {\textbf {Output}:} $\func{negamax}(u)$.\\  [1ex]
    {\textsc{negamax}($u$):}
    \begin{algorithmic}[1]
    \If {$|u.\attr{children}| = 0$}
      \State \Return $u.\attr{color} \cdot u.\attr{eval}$
    \Else
      \State $\var{value} := -\infty$
      \For{$v \in u.\attr{children}$}
        \State $\var{value} := \max(\var{value}, -\textsc{negamax}(v))$
      \EndFor
      \State \Return $\var{value}$
    \EndIf
    \Statex
    \Statex
    \Statex
    \Statex
    \Statex \phantom{$\bigcup$}
    \end{algorithmic}
    \end{algorithm}
\end{minipage}
\end{figure}
\algrenewcommand\algorithmicindent{1.4em}
%%%%%%%%%%%%%%%%%%%%%%%%%%%%%%%%%%%%%%%%%%%%%%%%%%%%%%%%%%%%%%%%%%%%%%%%%%%%

\section{Algorithm Pseudocode} \label{sec:algorithm-specifications}
This section introduces a selection of the game tree search algorithms that we have formally verified using Dafny. The classical \textsc{Minimax} algorithm (Algorithm~\ref{alg:minimax}) recursively evaluates a game tree to determine the optimal value at the root under perfect play. Each internal node loops over its children to compute either the minimum or maximum value, depending on the player to move.
Instead of handling separate cases for minimizing and maximizing nodes \textsc{Negamax} (Algorithm~\ref{alg:negamax}), is a reformulation of minimax that exploits symmetry between the players, using negation to unify both into a single recursive structure.
Negamax simplifies implementation, making it the de facto standard.
We further consider two negamax algorithms that combine depth-limited search, alpha-beta pruning, and transposition tables (TT), see Algorithm~\ref{alg:negamax-tt}. The first, which we refer to as \textsc{NegamaxTTW}, is a variant commonly described in community sources such as Wikipedia~(W), though its precise origin is unclear and likely influenced by work such as~\cite{Marsland86,Breuker1998}. The second, which we refer to as \textsc{NegamaxTTM}, is a related algorithm from Marsland~(M)~\cite{Marsland86}  that uses a different transposition table strategy. Special handling of the best move has been removed from \textsc{NegamaxTTM}.
Both algorithms consist of three distinct phases: a table lookup, a negamax search and a table update. In Section~\ref{sec:negamax-ttw-versus-negamax-ttm} the differences between them will be discussed.

\begin{algorithm}[tp]
\small
\caption{Transposition-table-based Negamax search variants.
Algorithm~\textsc{NegamaxTTW} follows Wikipedia~\cite{wiki:Negamax}, while
Algorithm~\textsc{NegamaxTTM} follows Marsland~\cite{Marsland86}.
Both algorithms take as input a node $u$, an alpha-beta window $(\alpha,\beta)$, a  search depth $d$, and a transposition table $T$.
Both algorithms are subdivided in three phases: Table Lookup, Negamax Search, and Table Update.
}
\label{alg:negamax-tt}

% ==================================================
\begin{algphase}{gray!25}{Header}
\textbf{function }\textsc{NegamaxTTW}($u, \alpha, \beta, d, T$):
\end{algphase}

\vspace{-5mm}

% ==================================================
\begin{algphase}{gray!10}{Phase 1: Transposition Table Lookup}

\State $\alpha_0 := \alpha$   \Comment{\textbf{Table Lookup}}
\If{$u \in T$} 
  \State $t := T[u]$
  \If{$t.\attr{depth} \ge d$}  
    \If{$t.\attr{flag} = \func{exact}$}
      \Return $t.\attr{value}$
    \ElsIf{$t.\attr{flag} = \func{lowerbound} \land t.\attr{value} \ge \beta$}
      \Return $t.\attr{value}$
    \ElsIf{$t.\attr{flag} = \func{upperbound} \land t.\attr{value} \le \alpha$}
      \Return $t.\attr{value}$
    \EndIf
  \EndIf
\EndIf

\end{algphase}

\vspace{-5mm}

% ==================================================
\begin{algphase}{white}{Phase 2: Negamax Search}

\If{$d = 0 \lor |u.\attr{children}| = 0$}
  \Return $u.\attr{color} \cdot u.\attr{eval}$      \Comment{\textbf{Negamax Search}}
\EndIf

\State $\var{value} := -\infty$
\For{$v \in u.\attr{children}$}
  \State $\var{value} := \max(\var{value},
    -\textsc{NegamaxTTW}(v,-\beta,-\alpha,d-1,T))$
  \State $\alpha := \max(\alpha,\var{value})$
  \If{$\alpha \ge \beta$}
    \textbf{break}
  \EndIf
\EndFor

\end{algphase}

\vspace{-5mm}

% ==================================================
\begin{algphase}{gray!20}{Phase 3: Transposition Table Update}

\If{$\var{value} \le \alpha_0$}
  $T[u] := (\var{value},d,\func{upperbound})$        \Comment{\textbf{Table Update}}
\ElsIf{$\var{value} \ge \beta$}
  $T[u] := (\var{value},d,\func{lowerbound})$
\Else
  \; $T[u] := (\var{value},d,\func{exact})$
\EndIf

\State \Return $\var{value}$

\end{algphase}

% ==================================================
\begin{algphase}{gray!25}{Header}
\textbf{function }\textsc{NegamaxTTM}($u, \alpha, \beta, \var{depth}, T$):
\end{algphase}

\vspace{-5mm}

% ==================================================
\begin{algphase}{gray!10}{Phase 1: Transposition Table Lookup}

\If{$u \in T$} \Comment{\textbf{Table Lookup}}
  \State $t := T[u]$
  \If{$t.\attr{depth} \geq d$}
    \If{$t.\attr{flag} = \func{exact}$}
        \Return $t.\attr{value}$
    \ElsIf{$t.\attr{flag} = \func{lowerbound}$}
        $\alpha := \max(\alpha, t.\attr{value})$
    \ElsIf{$t.\attr{flag} = \func{upperbound}$}
        $\beta := \min(\beta, t.\attr{value})$
    \EndIf
    \If{$\alpha \geq \beta$}
        \Return $t.\var{value}$
    \EndIf
  \EndIf
\EndIf

\end{algphase}

\vspace{-5mm}

% ==================================================
\begin{algphase}{white}{Phase 2: Negamax Search}

\If {$d = 0 \lor |u.\attr{children}| = 0$}
  \Return $u.\attr{color} \cdot u.\attr{eval}$  \Comment{\textbf{Negamax Search}}
\EndIf

\State $\var{value} := -\infty$
\For{$v \in u.\attr{children}$}
    \State $\var{value} := \max\bigl(\var{value}, -\textsc{NegamaxTTM}(v, -\beta, -\max(\alpha, \var{value}), d - 1, T)\bigr)$
    \If{$\var{value} \geq \beta$}
        \textbf{break}
    \EndIf 
\EndFor

\end{algphase}

\vspace{-5mm}

% ==================================================
\begin{algphase}{gray!20}{Phase 3: Transposition Table Update}

\State $\var{flag} := \func{exact}$  \Comment{\textbf{Table Update}}
\If{$\var{value} \leq \alpha$}
    $\var{flag} := \func{upperbound}$
\EndIf
\If{$\var{value} \geq \beta$}
    $\var{flag} := \func{lowerbound}$
\EndIf

\If {$u \notin T \lor T[u].depth \leq d$}
    $T[u] := (\var{value}, d, \var{flag})$
\EndIf    

\State \Return $\var{value}$

\end{algphase}

\end{algorithm}

%%%%%%%%%%%%%%%%%%%%%%%%%%%%%%%%%%%%%%%%%%%%%%%%%%%%

\section{Verification using Dafny} \label{sec:verification}

The modeling of data types in Dafny is straightforward, see Listing~\ref{lst:dafny-datatypes}. While in the pseudocode we used the value~$\infty$, in our Dafny modeling we use a bounded integer type, where \texttt{INFINITY} is an unspecified positive integer constant. This is also common in practical implementations. The verifier automatically checks that variables remain within the allowed domain. Note that recursive datatypes in Dafny, such as \texttt{Node}, are free from cycles by construction.
To verify correctness, we translated each algorithm from pseudocode into Dafny syntax. For example, \textsc{NegamaxTTW} in Algorithm~\ref{alg:negamax-tt} corresponds to Dafny Listing~\ref{lst:negamax-ttw}.

\begin{lstfloat}[t]
\begin{lstlisting}[basicstyle=\scriptsize\ttfamily]
  type bounded_int = x:int | -INFINITY <= x <= INFINITY
  datatype Color = White | Black
  datatype Node = Node_(eval:bounded_int, color:Color, children:seq<Node>)
  datatype Flag = Lowerbound | Upperbound | Exact
  datatype TableEntry = TableEntry_(depth:nat, value:bounded_int, flag:Flag)
  type TranspositionTable = map<Node, TableEntry>
\end{lstlisting}
\caption{The datatypes used in Dafny}
\label{lst:dafny-datatypes}
\end{lstfloat}

\subsection{Correctness of Transposition Table Search}
\begin{figure*}[tbh]
  \centering
  \includegraphics[width=\textwidth]{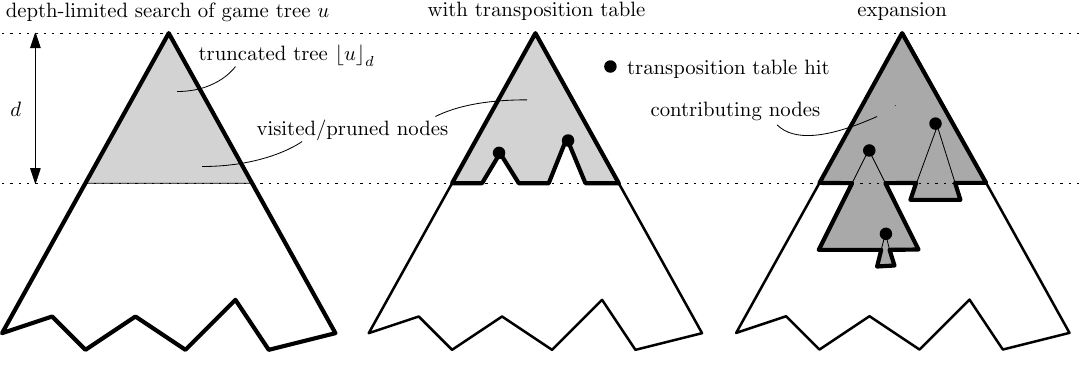}
  \caption{Schematic truncated game tree showing the visited/pruned nodes for depth-$d$ searching without (l) and with (m) transposition table. The expansion (r) with all nodes directly and indirectly contributing to the table-based search.}
  \label{fig:transposition-table-search}
\end{figure*}

For depth-limited transposition-table-based algorithms \textsc{NegamaxTT[W,M]}, the returned results may depend on values computed earlier in unrelated parts of the game tree, potentially at greater depth than required at the current node. Effectively, the algorithm explores an \emph{expansion} of the depth-$d$ truncated tree by greedily reusing deeper search results via the transposition table. Figure~\ref{fig:transposition-table-search}(r) illustrates this behavior: the primary search unfolds as a triangle-shaped subtree rooted at the current node. Upon a transposition table hit, the algorithm may reuse a  value previously computed in a deeper search of a different subtree.
So not only the nodes in the depth-$d$ truncated tree contribute to the search result, but also recursively the nodes in these (reused) subtrees. 
This motivates our notion of a \emph{witness}, a concrete expansion of the truncated tree that merges all such reused subtrees into a single game tree. 
Such an expansion $u'$ can be obtained from $u$ by removing all children of some nodes in $u$ at depth at least $d$.
Correctness is then characterized by the existence of a witness, an expansion whose minimax (or negamax) value equals the value returned by the algorithm.
To formalize this notion, we impose four conditions on a witness:
\begin{enumerate}
    \item The witness $u'$ is a subtree of the original tree $u$.
    %\item The truncated tree is included in the witness: $\truncate{u}{d} \subseteq u'$.
    \item The witness contains the complete depth-$d$ truncation of $u$: $\truncate{u}{d} = \truncate{u'}{d}$.
    \item For any node $z'$ in $u'$ corresponding to $z$ in $u$, either \emph{all} children of $z$ are included in $u'$, or \emph{none}.
    \item The algorithm’s returned value matches the negamax alpha-beta result on the witness $u'$, as defined in (\ref{eq:is-negamax-ab-result}).
\end{enumerate}
\noindent
The equality $\truncate{u}{d} = \truncate{u'}{d}$ in condition 2 is understood as isomorphism of the two truncated trees: they have identical branching structure and node evaluations within depth $d$. Condition 3 is crucial: it prevents a witness from including arbitrary subsets of a node’s children, which would produce subtrees that do not correspond to any valid minimax or negamax search.
Conditions 1-3 are formalized in the predicate \func{is-expansion}. Let $C = u.\attr{children}$ and $C' = u'.\attr{children}$:

\begin{eqnarray}
\label{eq:is-expansion}
\lefteqn{
\func{is-expansion}(u', u, d) \;=\;
u.\mathrm{eval} = u'.\mathrm{eval} \;\land\;
u.\mathrm{color} = u'.\mathrm{color} \;\land} \\[0.5em]
&& \begin{cases}
  \left( |C| = |C'| \;\land\;
  \forall\, i \in [0, |C|):\; \func{is-expansion}(C'_i, C_i, d - 1) \right),
  & \text{if } d > 0 \\[0.5em]
  \left( |C'| = 0 \;\lor\;
  \left( |C| = |C'| \;\land\;
  \forall\, i \in [0, |C|):\; \func{is-expansion}(C'_i, C_i, 0) \right) \right),
  & \text{if } d = 0
  \nonumber
\end{cases}
\end{eqnarray}
The first case ensures that the truncated tree $\truncate{u}{d}$ is included, while the second enforces
the “all-or-none children” requirement of Condition 3.

Correctness of the algorithm is then defined as:
\begin{equation}
\begin{array}{l}
\func{is-negamax-tt-result}(\var{result}, u, \alpha, \beta, \var{depth}) = \\[0.5em]
~~ \exists u':\; \func{is-expansion}(u', u, \var{depth}) \land 
\func{is-negamax-ab-result}(\var{result}, u', \alpha, \beta).
\end{array}
\label{eq:is-negamax-tt-result}
\end{equation}
Moreover, for each entry $t$ in the transposition table, a witness must exist justifying that entry:
\begin{eqnarray}
\label{eq:is-valid-table-entry}
\lefteqn{\func{is-valid-table-entry}(t, u) = \exists u': \func{is-expansion}(u', u, t.\attr{depth}) ~ \land} \nonumber \\
&& \quad \begin{cases}
    t.\attr{value} \geq \func{negamax}(u'), & \text{if } t.\attr{flag} = \textsf{upperbound} \\
    t.\attr{value} = \func{negamax}(u'),   & \text{if } t.\attr{flag} = \textsf{exact} \\
    t.\attr{value} \leq \func{negamax}(u'), & \text{if } t.\attr{flag} = \textsf{lowerbound}
\end{cases}
\end{eqnarray}
We adopt a permissive model of the transposition table. 
At every iteration of the algorithm, we allow the table's contents to change arbitrarily, as long as the entries satisfy~(\ref{eq:is-valid-table-entry}). This reflects practical implementations, where transposition tables are typically realized using hash tables: insertions may overwrite existing entries due to collisions and replacement strategies.

Importantly, this model does not restrict how a witness is obtained. Any witness satisfying Conditions 1-3 can be justified compositionally by valid transposition table entries, independently of a specific execution trace. Intuitively, the witness may be viewed as being assembled incrementally from shallow (e.g., one-ply) table lookups,
%each introducing a subtree justified by a previously stored entry,
but the correctness argument relies only on the existence of such a witness, not on its constructive derivation.

\subsection{Verification of \textsc{NegamaxTTW}}
We formally verified the Wikipedia variant of transposition-table-based negamax search, \textsc{NegamaxTTW}, using Dafny against the correctness criteria defined in Equations~(\ref{eq:is-negamax-tt-result}) and~(\ref{eq:is-valid-table-entry}). The Dafny development for this algorithm comprises approximately 850 lines of code, and verification completes in about 4 seconds on a standard workstation. The core algorithm is shown in Listing~\ref{lst:negamax-ttw}.

The verification effort involved several nontrivial challenges. First, correctness had to be expressed in terms of witnesses rather than concrete execution traces, requiring the specification of structural properties of expanded game trees. Second, suitable loop invariants were needed to relate intermediate search results, transposition table contents, and partially constructed witnesses. Third, the verification required reasoning about the incremental assembly of a witness tree that satisfies the expansion constraints introduced in Section~\ref{sec:definitions}.

Dafny proved effective for establishing many basic properties largely automatically. 
In particular, the \func{is-expansion} predicate admits a rich collection of lemmas that were relatively straightforward to verify, including reflexivity and transitivity, preservation under subtree inclusion, inclusion of the depth-truncated tree, and preservation of the expansion property when a subtree is replaced by a witness justified by a transposition table lookup.

At the same time, the verification process exposed limitations of automated reasoning in the presence of deeply nested quantifiers, which arise naturally when formalizing game trees, expansions, and transposition-table invariants. Without careful structuring, the verifier would often time out or exhaust memory. To address this, we systematically separated proof obligations and discharged them via explicit lemma invocations rather than inline reasoning, particularly at return points, loop exits, and invariant maintenance boundaries. For \textsc{NegamaxTTW}, this strategy resulted in five dedicated lemma calls in the implementation (Listing~\ref{lst:negamax-ttw}), covering early returns after table lookups, the $\var{depth}=0$ base case, loop maintenance and termination, and updates to the transposition table. This decomposition proved essential for scaling the verification to realistic algorithm variants.

\begin{lstfloat}[tp]
\begin{lstlisting}
class NegamaxAlgorithm
{
  var T: TranspositionTable

  method Negamax(u: Node, alpha0: bounded_int, beta0: bounded_int, depth:nat)
    returns (result: bounded_int)
    modifies this`T
    requires alpha0 < beta0
    requires turn_based()
    requires is_valid_table(T)
    ensures is_negamax_tt_result(result, u, alpha0, beta0, depth)
    ensures is_valid_table(T)
  {
    var alpha, beta := alpha0, beta0;
    if u in T.Keys
    {
      var t := T[u];
      if t.depth >= depth && ((t.flag == Lowerbound && t.value >= beta) ||
          (t.flag == Exact) || (t.flag == Upperbound && t.value <= alpha))
      {
        TableLookupReturnLemma(u, alpha, beta, depth, t, T);
        return t.value;
      }
    }
    if depth == 0 || |u.children| == 0
    {
      DepthZeroReturnLemma(u, depth, alpha0, beta0);
      return color(u) * u.eval;
    }
    var value: bounded_int := -INFINITY;
    for i := 0 to |u.children|
      invariant is_valid_table(T)
      invariant 0 <= i <= |u.children|
      invariant i == 0 ==> value == -INFINITY && alpha == alpha0
      invariant alpha0 <= alpha < beta0
      invariant value <= alpha0 ==> alpha == alpha0
      invariant alpha0 < value < beta0 ==> value == alpha
      invariant i > 0 ==> exists u': Node :: is_expansion(u', u, depth) &&
                   is_partial_negamax_ab_result(value, u', i, alpha0, beta0)
      invariant i == |u.children| ==>
              is_negamax_tt_result(value, u, alpha0, beta0, depth)
    {
      ghost var old_alpha := alpha;
      ghost var old_value := value;
      var v := u.children[i];
      var negamax_v := Negamax(v, -beta, -alpha, depth - 1);
      value := max(value, -negamax_v);
      alpha := max(alpha, value);
      if alpha >= beta
      {
        LoopBreakLemma(u, v, i, depth, alpha0, beta0, old_alpha, old_value,
                       alpha, value, negamax_v);
        break;
      }
      LoopMaintenanceLemma(u, v, i, depth, alpha0, beta0, old_alpha, old_value, alpha, value, negamax_v);
    }
    TableUpdateLemma(value, u, alpha0, beta0, depth, T);
    if value <= alpha0      {T := T[u:=TableEntry_(depth,value,Upperbound)];}
    else if alpha0 < value < beta0 {T:=T[u:=TableEntry_(depth,value,Exact)];}
    else if value >= beta0  {T := T[u:=TableEntry_(depth,value,Lowerbound)];}
    return value;
  }
}
\end{lstlisting}
\caption{Dafny verification of \textsc{NegamaxTTW}}
\label{lst:negamax-ttw}
\end{lstfloat}

%%%%%%%%%%%%%%%%%%%%%%%%%%%%%%%%%%%%%%%%%%%%%%%%%%%%%%%%%%%%%
\subsection{Verification of NegamaxTTM}
Our attempts to verify Marsland’s transposition table algorithm \textsc{NegamaxTTM} was ultimately unsuccessful.
During the Dafny proof, we encountered a specific configuration for which the postcondition of \func{is-negamax-tt-result} could not be established.
A closer analysis of this unprovable configuration led us to the discovery of a concrete counterexample, demonstrating that \textsc{NegamaxTTM} violates our witness-based correctness criterion.
The problematic case arises when the returned value~$x$ satisfies
\[
\alpha_0 < x \leq \alpha < \beta_0,
\]
where $\alpha_0$ and $\beta_0$ denote the original alpha-beta bounds of the call,
and $\alpha$ is the updated lower bound after a transposition table lookup
(e.g., $\alpha_0 = 0$, $x = 2$, $\alpha = 3$, $\beta_0 = 5$).

The configuration identified through formal verification turned out to be invaluable in our search for a counterexample.
It characterizes a situation in which a value returned from a transposition table lookup raises the lower bound~$\alpha$
beyond the eventual return value~$x$, while still remaining below the original upper bound~$\beta_0$.
This pattern suggests that a previously computed \emph{lower bound} may trigger premature pruning, thereby preventing the exploration of subtrees that would otherwise influence the minimax value.

\begin{figure*}[t]
\centering
\subfigure[Root call on $u$: depth~6, window $(0,5)$, returns 2 (exact)]{
\includegraphics[trim={1cm 1cm 1cm 1cm},clip,width=0.28\textwidth]{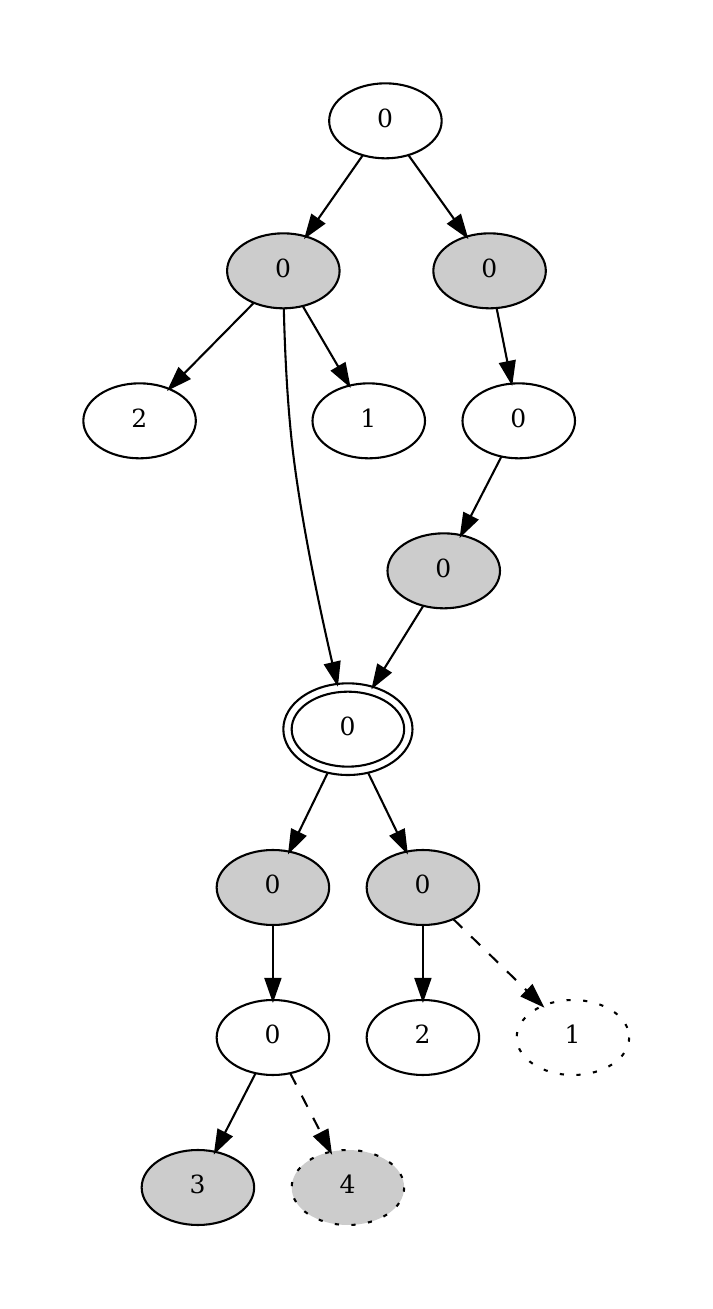}
\label{fig:counter-example-a}
}
\hspace{0.03\textwidth}
\subfigure[First call on $v$: depth~4, window $(0,2)$, returns 3 (lower bound)]{
\includegraphics[trim={1cm 1cm 1cm 1cm},clip,width=0.28\textwidth]{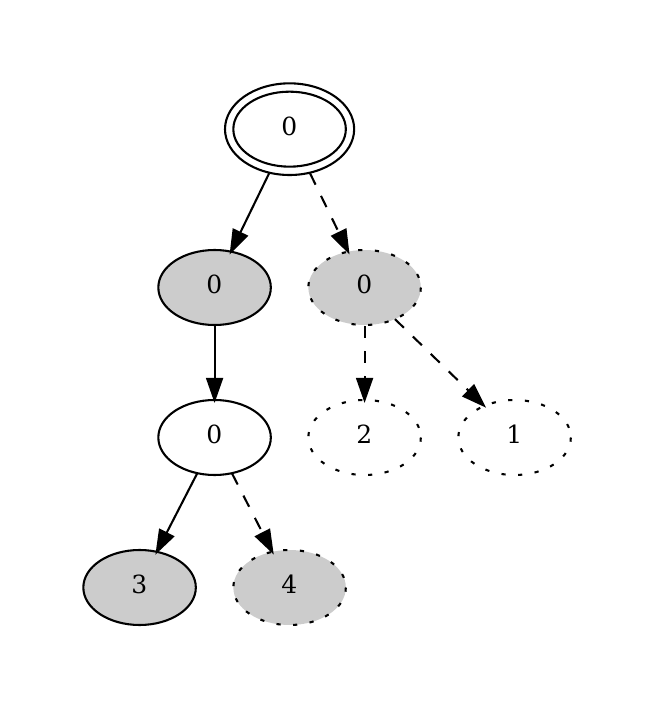}
\label{fig:counter-example-b}
}
\hspace{0.03\textwidth}
\subfigure[Second call on $v$: depth~2, window $(0,5)$, reuses (b)'s result and returns 2 (exact)]{
\includegraphics[trim={1cm 1cm 1cm 1cm},clip,width=0.28\textwidth]{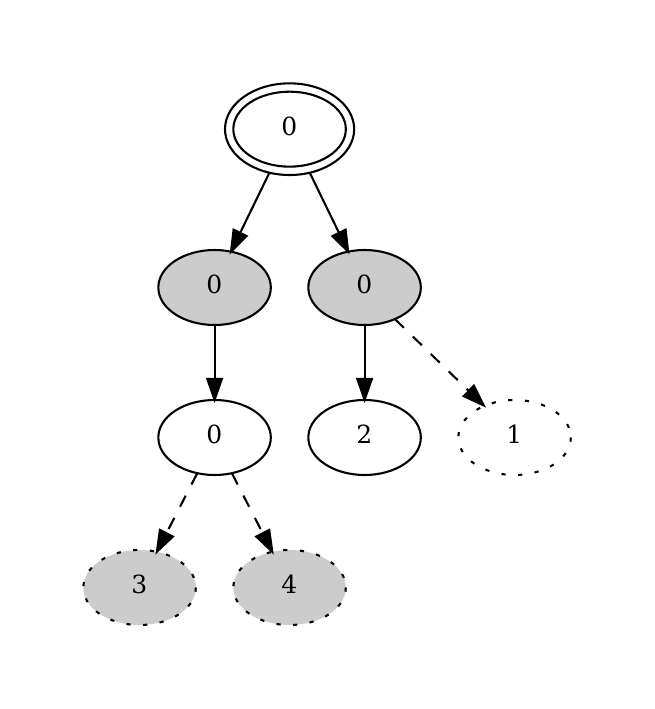}
\label{fig:counter-example-c}
}
\caption{Counterexample to \textsc{NegamaxTTM}. Node $v$ (with double border) is searched twice. Call (b) stores a lower bound of 3 in the transposition table. Call (c) reuses this result in a shallower search with a wider window, leading to an unexpected return value. Dotted nodes are not visited during search.}
\label{fig:counter-example}
\end{figure*}

In \textsc{NegamaxTTM}, such a situation can arise only when the lower bound~$\alpha$ is increased by a transposition table lookup
that yields a \emph{lower bound} computed at a greater or equal search depth.
Guided by this observation, we constructed a concrete instance realizing this mechanism by considering the window $(\alpha,\beta) = (0,5)$, together with the transposition table
\[
T = \{ v \mapsto (\var{value}=3, \var{depth}=4, \var{flag}=\func{LB}) \},
\]
followed by a search of the same node~$v$ at depth~2 that returns a value $x=2$.
After substantial manual analysis, this led to the hand-crafted counterexample shown in Figure~\ref{fig:counter-example}(c).

Applied to this example, \textsc{NegamaxTTM} indeed returns the value~2, as shown in Equation~\ref{eq:counterexample}(c).
This outcome is problematic for two independent reasons.
First, it violates standard minimax intuition: the parent of the node with value~2 is a minimizing node, and under optimal play at depth~2 it should instead select the branch with value~1.
Second, and more importantly for our formal development, there exists no witness tree satisfying the criterion~(\ref{eq:is-negamax-tt-result}) that justifies the return value~2.

Indeed, there are only two expansions of $\truncate{u}{2}$ that satisfy the $\func{is-expansion}$ predicate, namely $\truncate{u}{2}$ itself and the full tree~$u$.
The corresponding negamax alpha-beta results for these expansions are~1 and~4, respectively; the value~2 cannot be obtained from any admissible witness.
Crucially, the node with value~1 is never visited during the search.
This is a direct consequence of the lower-bound value~3 stored in the transposition table, which shrinks the alpha--beta window to $(3,5)$ during the table-lookup phase.
When the node with value~2 is subsequently evaluated, the loop’s break condition is triggered, and no further children are explored.

At first glance, this counterexample might appear unconvincing, since it assumes a specific non-empty transposition table at the start of the search.
To show that this situation can arise in a complete execution, we therefore extend the example to a run starting from an empty table, as shown in Figure~\ref{fig:counter-example}(a).
The algorithm is invoked at the root node~$u$ with window $(0,5)$, depth~6, and an initially empty transposition table.
This call generates two recursive searches of node~$v$ (indicated by a double border), and no table entries are removed during the search.

The resulting calls are summarized in Equation~\ref{eq:counterexample}.
In particular, call~(b) stores a lower bound of~3 for~$v$ at depth~4 in the table.
This entry is then reused in call~(c), which searches~$v$ at depth~2 with a wider window.
The reuse of this lower bound again causes premature pruning and leads to the return of the problematic value~2.
Also in this extended execution, no witness exists that justifies the returned result.

\begin{alignat}{3}
  \label{eq:counterexample}
  \text{~~(a)~~} &\textsc{NegamaxTTM}(u, \alpha=0, \beta=5, \var{depth}=6, T=\emptyset) &\quad &= 2 &\quad& (\func{exact}) \\
  \text{~~(b)~~} &\textsc{NegamaxTTM}(v, \alpha=0, \beta=2, \var{depth}=4, T)            &\quad &= 3 &\quad& (\func{lowerbound}) \notag \\
  \text{~~(c)~~} &\textsc{NegamaxTTM}(v, \alpha=0, \beta=5, \var{depth}=2, T)            &\quad &= 2 &\quad& (\func{upperbound}) \notag
\end{alignat}
The root cause of this failure is the reuse of a lower bound computed under a narrower alpha-beta window $(0, 2)$ in a subsequent search with wider window $(0, 5)$. This reuse leads to pruning of a node that would otherwise yield a better minimax value.

The counterexample shows that \textsc{NegamaxTTM} violates our witness-based correctness notion: no witness exists that justifies the returned value. Moreover, the returned value exhibits behavior that is difficult to reconcile with standard minimax semantics. In particular, a minimizing node selects a child with value~2 even though another child with value~1 lies within the search horizon and is never explored.
These observations do not imply that \textsc{NegamaxTTM} is incorrect as a practical search algorithm. However, it remains an open question whether a useful correctness criterion exists.

\subsection{Execution of the Minimal Counterexample}
To understand why \textsc{NegamaxTTM} violates our correctness criterion while \textsc{NegamaxTTW} satisfies it, we examine their concrete executions on the minimal counterexample of Figure~\ref{fig:counter-example-c}. Both algorithms are called with identical inputs:
\[
\texttt{NegamaxTT[W,M]}(v, \alpha=0, \beta=5, \mathit{depth}=2, T),
\]
where $T = \{ v \mapsto (\mathit{value}=3, \mathit{depth}=4, \mathit{flag}=\textsf{LB}) \}$.

\begin{figure}[tb]
\centering
\includegraphics[width=\textwidth]{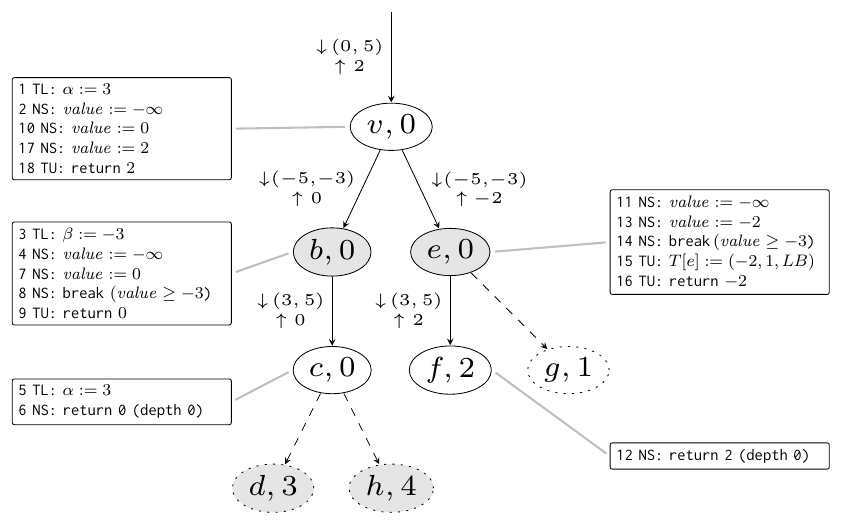}
\caption{Execution trace of \textsc{NegamaxTTM} on the minimal counterexample from Figure~\ref{fig:counter-example-c}, with transposition table $\{ v \mapsto (\mathit{value}=3, \mathit{depth}=4, \mathit{flag}=\textsf{LB}) \}$. Nodes contain an identifier and an evaluation value. Rectangular blocks show statements executed at each node, labeled by algorithmic phase (TL/NS/TU) and global execution order. Edge labels show the alpha-beta interval passed downward ($\downarrow$) and the value returned upward ($\uparrow$).}
\label{fig:execution-ttm}
\end{figure}

\subsubsection{NegamaxTTM}
Figure~\ref{fig:execution-ttm} shows the recursive depth-first search of \textsc{NegamaxTTM} through nodes $v$, $b$, $c$, $e$, and $f$. For each node, a text block lists the relevant statements executed during its search, prefixed with their global execution order. The statements are labeled by phase: \textbf{TL} (Table Lookup), \textbf{NS} (Negamax Search), and \textbf{TU} (Table Update). Edges are labeled with two pieces of information:
\begin{itemize}
\item $\downarrow (\alpha,\beta)$: the alpha-beta window passed to the child, after the standard negation and swapping of bounds ($-\beta, -\alpha$),
\item $\uparrow \mathit{value}$: the value returned by the child.
\end{itemize}
Dashed edges and nodes (e.g., $d$, $g$, $h$) are not visited during this search.
\noindent
The key steps that lead to the problematic return value $2$ are:
\begin{enumerate}
\item At the root $v$, the table lookup increases $\alpha$ from $0$ to $3$ (step~1). This narrowing of the window is caused by the stored lower bound for $v$.
\item When searching child $b$, the window becomes $(-5,-3)$ after negation. Its child $c$ returns its own value $0$ (step~6), since it is searched at depth $0$. The search of $b$ propagates this value $0$ (step~9) to the root $v$ (step~10).
\item Child $e$ is searched with window $(-5,-3)$. Its leaf child $f$ returns $2$, which becomes $-2$ after negation (step~13). Since $-2 \ge -3$, a beta-cutoff occurs (step~14), skipping the better child $g$ with value $1$. The value $-2$ is stored in the TT (step~15) and propagated back to the root $v$ (step~16).
After negation (step~17) the final return value becomes $2$.
\end{enumerate}
The stored lower bound $3$ for $v$ thus causes a premature cutoff at $e$. Consequently, the algorithm never visits the subtree that would reveal the better minimax value $1$ available at depth~2. The return value $2$ cannot be justified by any witness expansion of $\truncate{v}{2}$, confirming the violation of our correctness criterion.

\begin{figure}[tbh]
\centering
\includegraphics[width=\textwidth]{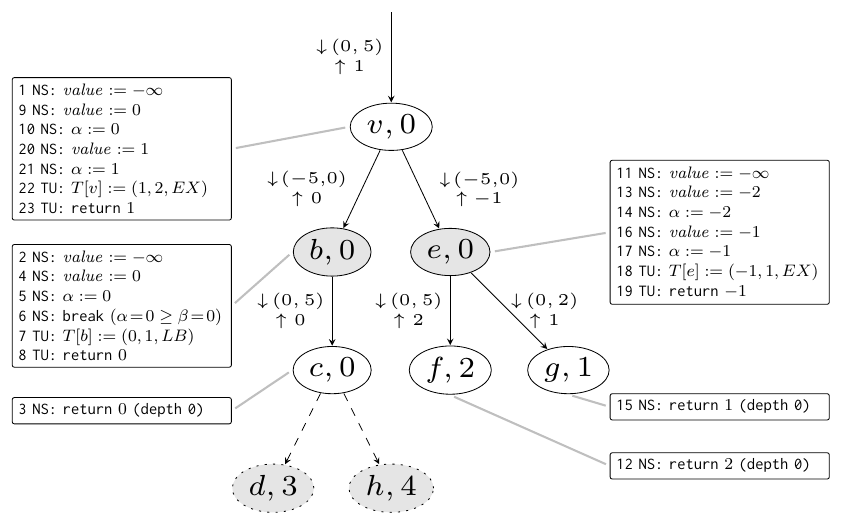}
\caption{Execution trace of \textsc{NegamaxTTW} on the minimal counterexample from Figure~\ref{fig:counter-example-c}, with transposition table $\{ v \mapsto (\mathit{value}=3, \mathit{depth}=4, \mathit{flag}=\textsf{LB}) \}$.}
\label{fig:execution-ttw}
\end{figure}

\subsubsection{NegamaxTTW}
Figure~\ref{fig:execution-ttw} illustrates the execution of \textsc{NegamaxTTW}. The table lookup at $v$ also finds the stored lower bound $3$, but the check $t.\attr{value} \ge \beta$ (i.e., $3 \ge 5$) fails, so the lookup does \emph{not} cause an early return. Consequently, the search proceeds with the original window $(0,5)$, explores all relevant children, and returns the expected negamax value $1$. The stored entry $3$ is effectively ignored because it does not guarantee a cutoff in the current, wider window.

\subsubsection{Comparison}
The two traces highlight the semantic difference between the table-lookup strategies. \textsc{NegamaxTTM} uses stored bounds to narrow the search window, which can lead to incorrect pruning when a bound computed under a narrow window is reused in a wider one. \textsc{NegamaxTTW} only accepts a stored value if it immediately guarantees a cutoff; otherwise it discards the entry and performs a full search. This more conservative policy preserves the existence of a witness tree, explaining why \textsc{NegamaxTTW} satisfies our correctness criterion while \textsc{NegamaxTTM} does not.

%%%%%%%%%%%%%%%%%%%%%%%%%%%%%%%%%%%%%%%%%%%%%%%%%%%%%%%%
\subsection{Differences between NegamaxTTW and NegamaxTTM}
\label{sec:negamax-ttw-versus-negamax-ttm}
The algorithms \textsc{NegamaxTTW} and \textsc{NegamaxTTM} in Algorithm~\ref{alg:negamax-tt} both consist of three conceptual phases: a table lookup, a negamax search, and a table update. While structurally similar, the two algorithms differ in each of these phases:
\begin{itemize}
    \item \textbf{Table Lookup:}
    \textsc{NegamaxTTM} uses upper and lower bounds from the transposition table to narrow the alpha-beta window, while \textsc{NegamaxTTW} \\ terminates the search immediately when a matching entry is found.

    \item \textbf{Negamax Search:}
    \textsc{NegamaxTTM} follows Fishburn’s fail-soft negamax scheme~\cite{Fishburn83}, which differs in how intermediate
    bounds are propagated during recursion.

    \item \textbf{Table Update:}
    \textsc{NegamaxTTM} uses the current value of $\alpha$ to classify search results, while \textsc{NegamaxTTW} uses the initial value of $\alpha$ for that. In addition, \textsc{NegamaxTTM} preserves existing table entries when their recorded search depth exceeds that of the newly computed result.
\end{itemize}
We analyzed the impact of each of these differences on correctness with respect to our witness-based criterion. The verification attempt for \textsc{NegamaxTTM} shows that the deviation in the \emph{Table Lookup} phase is decisive: reusing bounds to shrink the search window can lead to results for which no admissible witness expansion exists, and therefore violates our correctness notion.

The remaining differences were studied in isolation by incorporating them into \textsc{NegamaxTTW} and re-verifying the resulting variants in Dafny. The additional depth condition in the \emph{Table Update} phase does not affect correctness, since our transposition table model constrains only the validity of stored entries, not their presence or replacement policy.

Similarly, adopting Fishburn’s fail-soft negamax variant in the \emph{Negamax Search} phase does not compromise correctness. Proving this required adaptations to the loop maintenance and loop break lemmas, but no changes to the underlying correctness argument.

In contrast, classifying results against the \emph{current} value of $\alpha$ rather than the initial one turned out to be more subtle. Under this modification, the three cases in the table-update are no longer mutually exclusive. Correctness could only be re-established by imposing a different ordering on these cases, ensuring that the correct flag (lower bound, exact or upper bound) is assigned to the result.
%%%%%%%%%%%%%%%%%%%%%%%%%%%%%%%%%%%%%%%%%%%%%%%%%%%%%%%%

\section{Results}
Our first contribution is the introduction of a witness-based correctness criterion for depth-limited game tree search with transposition tables. This notion formalizes the intuitive requirement that the returned value must correspond to the exact evaluation of some fully explored subtree (the witness).

As a second contribution, we applied this correctness notion to two representative algorithms. We successfully verified the \textsc{NegamaxTTW} algorithm in Dafny, providing, to our knowledge, the first formal proof of correctness for a depth-limited transposition-table-based negamax search with alpha-beta pruning. In contrast, the algorithm by Marsland~\cite{Marsland86} (with best move tracking omitted) was shown to violate the criterion, based on a concrete counterexample. 
Whether this outcome reflects a shortcoming in Marsland's algorithm or highlights the need for a different correctness notion remains an open question.

Beyond these case studies, we developed and verified over 15 variations of minimax and negamax algorithms, ranging from basic recursive definitions to depth-limited and alpha-beta optimized versions. All verification artifacts, including Dafny proofs and Python implementations with randomized testing, are available on GitHub~\cite{github-minimax}. The most complex case, \textsc{NegamaxTTW}, verifies in approximately 4 seconds with Dafny 4.10.0.

\section{Conclusions}
This work illustrates the value of formal verification in reasoning about correctness properties of game tree search algorithms, particularly when optimizations such as transposition tables are employed.

Our primary contribution is a general and intuitive witness-based correctness criterion for depth-limited search using transposition tables, applicable both with and without alpha-beta pruning. We demonstrate that the proposed criterion can be used to verify a variant found on Wikipedia (which we call \textsc{NegamaxTTW}), whereas it does not hold for a closely related variant by Marsland~\cite{Marsland86} (\textsc{NegamaxTTM}).
This contrast highlights how subtle algorithmic differences can invalidate seemingly plausible correctness arguments, underscoring the necessity of precise semantic models when verifying optimized search techniques.
More broadly, our work illustrates Dafny’s suitability for verifying recursive, stateful algorithms with nontrivial invariants, as commonly found in AI search for games.

Several directions for future work emerge from this study. A natural next step is to extend the witness-based framework to other search algorithms that rely intrinsically on the persistence of transposition table state, such as $SSS^\ast$ \cite{Stockman79,PlaatSPB96} and MTD($f$) \cite{PlaatSPB96}. Another goal is to find a relaxed correctness notion for \textsc{NegamaxTTM}, whose behavior is not characterized by our current witness-based criterion.
Recent advances in quantum program verification (e.g.,~\cite{Ying11}) further suggest a pathway toward extending formal, semantics-based correctness reasoning to quantum variants of game tree search.

\bibliographystyle{splncs04}
\bibliography{main}

\end{document}